\pdfoutput=1
\documentclass[11pt]{article}

\usepackage{acl}

\usepackage{times}
\usepackage{latexsym}
\usepackage{graphicx}

\usepackage[T1]{fontenc}
\usepackage[utf8]{inputenc}

\usepackage{microtype}
\usepackage{multirow}
\usepackage{multicol}

\title{Variation and Instability in Dialect-Based Embedding Spaces}

\author{Jonathan Dunn \\
  Department of Linguistics and \\
  New Zealand Institute for Language, Brain and Behaviour \\
  University of Canterbury \\
  Christchurch, New Zealand \\
  \texttt{jonathan.dunn@canterbury.ac.nz} \\}

\begin{document}
\maketitle
\begin{abstract}
This paper measures variation in embedding spaces which have been trained on different regional varieties of English while controlling for instability in the embeddings. While previous work has shown that it is possible to distinguish between similar varieties of a language, this paper experiments with two follow-up questions: First, does the variety represented in the training data systematically influence the resulting embedding space after training? This paper shows that differences in embeddings across varieties are significantly higher than baseline instability. Second, is such dialect-based variation spread equally throughout the lexicon? This paper shows that specific parts of the lexicon are particularly subject to variation. Taken together, these experiments confirm that embedding spaces are significantly influenced by the dialect represented in the training data. This finding implies that there is semantic variation across dialects, in addition to previously-studied lexical and syntactic variation.
\end{abstract}

\section{Dialects and Embedding Spaces}

This paper investigates the degree to which embedding spaces are subject to variation according to the regional dialect or variety that is represented by the training data. The experiments train character-based skip-gram embeddings on gigaword corpora representing four regional dialects of English (North America, Europe, Africa, and South Asia). While there is a robust tradition of discriminative modelling of dialects and varieties within \textsc{nlp} \cite{zampieri-etal-2017-findings, zampieri-etal-2018-language, zampieri-etal-2019-report, gaman-etal-2020-report, chakravarthi-etal-2021-findings-vardial, aepli-etal-2022-findings}, there has been much less work on the influence which the dialectal composition of the training data (upstream) has on embedding spaces after training (downstream). 

\begin{figure*}[t]
\centering
\includegraphics[width = 380pt]{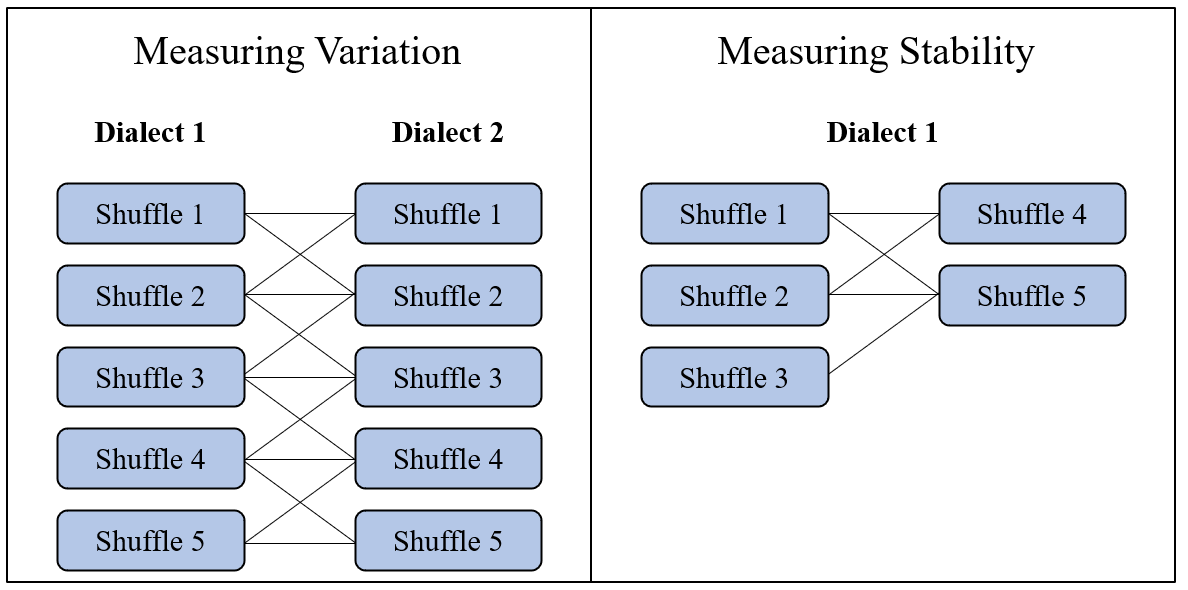}
\caption{Overview of Comparison Methodology: Variation between dialects is estimated by sampling ten unique pairs of embeddings, where each embedding represents a shuffled version of a dialect-specific corpus. The baseline instability is estimated by sampling ten unique pairs of embeddings from different shuffled versions of a single dialect-specific corpus. Non-geographic factors like time period and random seed are held constant.}
\label{fig_method}
\end{figure*}

The basic idea in this paper is to train five iterations of character-based skip-gram embeddings on dialect-specific corpora in order to measure both variation (across dialects) and instability (within dialects); this is visualized in Figure \ref{fig_method}. In order to find out whether specific parts of the lexicon are especially influenced by the dialect represented in the training data, the lexicon used for comparing embedding spaces is annotated for frequency, concreteness, part-of-speech, semantic domain, and age-of-acquisition.

If the specific dialect represented in the training corpus has no influence on embedding spaces, then variation across regions will be the same as variation within regions. In other words, we must control for instability (operationalized as variation across embeddings from the same dialect) to avoid false positives. However, if the dialect represented in the training data does have an influence on embedding spaces after training, then there will be a clear distinction between variation across dialects and instability within dialects.

The contribution of this paper is to show (i) that dialectal variation in character-based embedding spaces is significantly stronger than the noise caused by background instability and (ii) that this variation remains concentrated in certain parts of the lexicon. To accomplish this, we model the impact of dialect-specific training corpora on embeddings by controlling for background instability and organizing the experiments around the lexical attributes of frequency, concreteness, part-of-speech, semantic domain, and age-of-acquisition.

We begin by reviewing related work on dialectal variation and embedding stability (Section 2), before describing the main experimental questions (Section 3), the data (Section 4), and the methods (Section 5). We then compare variation within and between dialect-specific embeddings (Section 6) before modelling the influence of lexical factors on such dialectal variation (Section 7). Taken together, these experiments confirm that regional dialect or variety has a significant influence on embedding spaces that far exceeds baseline instability.

\section{Related Work}

This section discusses previous work in which models trained on data from different varieties (upstream) become significantly different after training (downstream). It also presents previous work on instability in embedding spaces and the factors that influence such instability.

\begin{table*}[t]
\centering
\begin{tabular}{|lll|c|c|}
\hline
\textbf{Circle} & \textbf{Region} & \textbf{Country} & \textbf{N. Words, Web} & \textbf{N. Words, Tweets} \\
\hline
\multirow{2}{*}{Inner-Circle} & \multirow{2}{*}{North American} & Canada & 250 mil & 250 mil \\
& & United States & 250 mil & 250 mil \\
\hline
\multirow{2}{*}{Inner-Circle} & \multirow{2}{*}{European} & Ireland & 250 mil & 250 mil \\
& & United Kingdom & 250 mil & 250 mil \\
\hline
\hline
\multirow{4}{*}{Outer-Circle} & \multirow{4}{*}{African} & Nigeria & 262 mil & 100 mil \\
~ & ~ & Kenya & 1 mil & 100 mil \\
~ & ~ & Gabon & 100 mil & 100 mil \\
~ & ~ & Uganda & 37 mil & 100 mil \\
~ & ~ & Mali & 100 mil & 100 mil \\
\hline
\multirow{2}{*}{Outer-Circle} & \multirow{2}{*}{South Asian} & India & 250 mil & 250 mil \\
~ & ~ & Pakistan & 250 mil & 250 mil \\
\hline
  \end{tabular}
  \caption{Source of Data by Region, Country, and Register}
  \label{tab:1}
\end{table*}

\textbf{Geography and Dialects}. The creation of large geo-referenced corpora has made it possible to model variation across dialects, where unique locations represent unique dialect regions. Previous work has described geo-referenced corpora derived from web pages and social media \cite{Davies2015, Dunn2020}. Other work has evaluated the degree to which such corpora represent dialectal patterns found using more traditional methods \cite{Cook2017, Grieve2019}, and the degree to which these corpora capture population movements triggered by events like the \textsc{covid-19} pandemic \cite{Dunn2020b}. Further work has shown that geographic corpora from distinct sources largely agree on their representation of national dialects \cite{Dunn2021}. Building on these corpora, recent work has modelled both lexical variation  \cite{Wieling2011, Donoso2017, Rahimi2017a} and syntactic variation \cite{Dunn2018, Dunn2019a, DunnWong2022} in English as well as in other languages \cite{Dunn2019}. 

To what degree does dialectal variation influence semantic representations like skip-gram embeddings in addition to lexical and syntactic features? Previous work has shown that there is a significant difference between generic web-based embeddings and web-based embeddings trained using corpora sampled to represent actual population distributions; this difference was observed across 50 languages \cite{Dunn2020a}. While these previous results lead us to expect dialectal variation across embeddings, there are two remaining questions: First, to what degree is this variation caused by dialectal differences as opposed to random instability? Second, is dialectal variation spread equally across the lexicon, equally influencing nouns and verbs, abstract and concrete, frequent and infrequent words?

\textbf{Instability in Embeddings}. A related line of work focuses on sources of instability in embedding spaces. It has been shown that many embeddings are subject to random fluctuation across different cycles of shuffling and retraining \cite{Hellrich2019}. Such instability has been investigated using word similarities \cite{Antoniak2018}, showing that smaller corpora are subject to greater instability. In this line of work, two embeddings are compared by measuring the overlap in nearest neighbors for a target vocabulary. It has been shown, for example, that even high-frequency words can be unstable \cite{Wendlandt2018} and that instability is related to properties of a language like the amount of inflectional morphology \cite{Burdick2021}. Other work has focused on the impact of time on embeddings, with variation leading to change \cite{Cassani2021}.

Other recent work has shown that register variation \cite{Biber2012} has a significant impact on embedding similarity across a diverse range of languages \cite{Dunn2022}.  This general approach to comparing embedding spaces focuses on aligned vocabulary (using nearest neighbors) rather than aligned embeddings because of instability in such alignment methods themselves \cite{gonen-etal-2020-simple}. As shown by this previous work, the comparison of nearest neighbors provides a robust method for detecting variation across embedding spaces.

This work on instability in embeddings is important because we need to distinguish between (i) variation across dialects and (ii) random fluctuations in embedding representations themselves. In other words, given the finding that embeddings trained on corpora representing different dialects are significantly different, how much of this is noise caused by random instability?

\section{Experimental Questions}

This paper focuses on two questions: First, are there significant differences in embeddings trained from corpora representing different dialects when accounting for baseline instability in the embeddings? Second, if so, are these dialectal differences specific to a certain part of the vocabulary, such as words belonging to a specific semantic domain?

The basic idea here is to compile four gigaword corpora representing English as used in North America, Europe, Africa, and South Asia. These areas represent different dialect regions. For example, while there are smaller differences between American English and Canadian English, these two dialects are more similar to one another than to other national dialects like Irish English. For example, work on syntactic variation has shown that American and Canadian English, at least in digital contexts, are closely related while UK and Irish English form a separate closely related pair \cite{Dunn2019}. Based on the distribution of errors within a confusion matrix, other work has shown that Indian and Pakistani English are likewise more similar to one another than to other dialects \cite{Dunn2018}.

The dialects included represent both inner-circle and outer-circle varieties. The concept of inner-circle vs outer-circle is based on the historical stages of European colonization \cite{Kachru1982}. This distinction within the World Englishes paradigm is meant to capture the perceived prestige differences of these dialects rather than to make a distinction between dialects and varieties as linguistic objects. For example, inner-circle populations tend to have a higher socio-economic status and better access to digital technologies, leading to their status as prestige varieties. Both groups can be considered dialects. In some cases speakers of outer-circle varieties could be considered second-language learners; however, regardless of a distinction between native and non-native speakers, the production found in outer-circle varieties remains robust and predictable over time. Thus, we treat both inner-circle and outer-circle varieties as dialects of equal standing but maintain the terminology from the World Englishes paradigm in order to provide a bridge to work in sociolinguistics.

We first train embeddings on each dialect-specific corpus and then measure variation across a lexicon that is annotated for concreteness, age-of-acquisition, semantic domain, part-of-speech, and frequency. We train five sets of embeddings for each dialect-specific corpus, each based on a random reshuffling of the corpus. This allows us to measure the difference between variation (across dialects) and baseline instability (within dialects).

We work with skip-gram embeddings (\textsc{sgns}: \citealt{Mikolov2013b}) as implemented in the fastText framework \cite{Bojanowski2017}. In particular, we use the skip-gram variant with negative sampling ($n=50$) trained for 20 epochs with a learning rate of 0.05 and 100 dimensions. The character n-gram sizes range from 3 to 6, with a maximum of 2 million n-gram buckets allowed. Because previous work has shown that different random seeds can cause instability \cite{gonen-etal-2020-simple}, we control for such instability by using the same random seed for each set of embeddings. Thus, variation caused by random seed and by training parameters is taken into account in this experimental set-up.

Several considerations support the use of non-contextual skip-gram embeddings for these experiments. In the first case, the focus here is on semantic variation rather than lexical or syntactic structures and the long-distance co-occurrences captured by the skip-gram task are taken as better representations for such semantic variation. In the second case, the inclusion of low-resource dialects like African English means that the amount of training data available is limited and insufficient for training robust contextual embeddings. Given the dual goals of focusing on semantics while also including low-resource dialects, skip-grams provide the most practical type of embedding for answering these particular experimental questions.

\section{Data}

The data used here represents different geographic locations which, in turn, represent different dialects. The data itself is drawn from two registers, web pages and tweets, both derived from the \textit{Corpus of Global Language Use} \cite{Dunn2020}. The experiments train character-based embeddings for these four different regional dialects, as shown in Table \ref{tab:1}. Each corpus contains 1 billion words, equally divided between registers (web pages and tweets). Thus, for example, the inner-circle North American corpus contains 500 million words of tweets, equally divided between Canada and the United States. The African web corpus has additional constraints because there is less data per country. As shown in Table \ref{tab:1} this corpus combines five countries into a single regional data set. The even split between web pages and tweets is maintained.

\section{Methods}

For each regional variety of English, we train embeddings using the fastText framework with the parameter settings described above. Previous work has shown that this family of embeddings can be unstable; in this context, \textit{instability} means that the same training corpus could result in multiple sets of nearest neighbors over different iterations \cite{Hellrich2019}. We control for this by randomly shuffling each corpus and retraining the embeddings five times. Because all comparisons are between two sets of embeddings, we thus obtain ten observations (unique comparisons) to represent each condition, as visualized in Figure \ref{fig_method}. We use the same random seed and the same parameters across all sets of embeddings to control for other sources of variation.

\textbf{Vocabulary Features}. The vocabulary for the embedding space is derived from semantic and psycholinguistic resources that provide categorizations for specific lexical items. This source of vocabulary allows us to compare stability and variation across different sub-sets of the lexicon.

\begin{table}[h]
\centering
\begin{tabular}{|lr|lr|}
\hline
\textbf{Concreteness} & \textbf{N.} & \textbf{POS} & \textbf{N.}\\
\hline
1.0 to 2.0 & 2,426 & Adjective & 4,130 \\
2.0 to 3.0 & 5,619 & Adverb & 189 \\
3.0 to 4.0 & 4,167 & Name & 139 \\
4.0 to 5.0 & 4,599 & Noun & 9,827 \\
- & - & Verb & 2,322 \\
- & - & Other & 205 \\
\hline
\textbf{Total} & \textbf{16,812} & \textbf{Total} & \textbf{16,812} \\
\hline
  \end{tabular}
  \caption{Distribution of Vocabulary Items Across Concreteness Categories and Parts-of-Speech}
  \label{tab:3}
\end{table}

\begin{table}[h]
\centering
\begin{tabular}{|lrcc|}
\hline
\textbf{Category} & \textbf{N.} & \textbf{Conc} & \textbf{AoA} \\
\hline
General \& Abstract & 2,384 & 2.4 & 10.5 \\
Body \& Individual & 1,268 & 3.8 & 9.8 \\
Arts \& Crafts & 114 & 3.8 & 9.8 \\
Emotion & 765 & 2.3 & 9.9 \\
Food \& Farming & 586 & 4.2 & 8.6 \\
Government \& Public & 761 & 2.9 & 10.9 \\
Housing \& Home & 336 & 4.2 & 8.7 \\
Money \& Commerce & 531 & 3.2 & 10.5 \\
Entertainment & 459 & 3.9 & 8.7 \\
Life \& Living Things & 594 & 4.3 & 8.3 \\
Movement \& Travel & 897 & 3.5 & 9.1 \\
Numbers \& Measures & 795 & 2.8 & 9.7 \\
Materials \& Objects & 1,806 & 3.7 & 9.0 \\
Education & 118 & 3.3 & 9.8 \\
Communication & 943 & 3.2 & 9.9 \\
Social Actions & 1,959 & 2.7 & 10.2 \\
Time & 474 & 2.7 & 9.2 \\
World \& Environ. & 298 & 3.9 & 8.6 \\
Psychological & 1,255 & 2.4 & 9.8 \\
Science \& Tech & 161 & 3.3 & 11.4 \\
Names \& Grammar & 307 & 2.9 & 7.5 \\
\hline
\textbf{Total} & \textbf{16,812} & \textbf{3.1} & \textbf{9.7} \\
\hline
  \end{tabular}
  \caption{Distribution of Vocabulary Items Across Semantic Domains with Concreteness and Age-of-Acquisition Information for Each Domain}
  \label{tab:4}
\end{table}

The first source of lexical annotations is a participant-based study of concreteness \cite{Brysbaert2014}. This source provides concreteness ratings between 1 and 5 for each lexical item, with higher values reflecting more concrete and lower values reflecting more abstract judgements from participants. This source also provides the most common part-of-speech for each lexical item. The distribution of the vocabulary across concreteness ratings and parts-of-speech is shown in Table \ref{tab:3}. An example of an abstract word (1.0 to 2.0) is \textit{belief}; less abstract (2.0 to 3.0) is \textit{famished}; more concrete (3.0 to 4.0) is \textit{galaxy}; and most concrete (4.0 to 5.0) is \textit{fire}. Within parts-of-speech, most words are categorized as adjectives, adverbs, nouns, or verbs.

\begin{table}[h]
\centering
\begin{tabular}{|r|c|ccc|}
\hline
\multirow{2}{*}{\textbf{Word}} & \textbf{Stability} & \multicolumn{3}{|c|}{\textbf{Overlap}} \\
~ & \textbf{\textsc{na}} & \textbf{\textsc{eu}} & \textbf{\textsc{af}} & \textbf{\textsc{sa}} \\
\hline
\textit{shag} & 0.53 & 0.00 & 0.00 & 0.03 \\
\textit{daft} & 0.59 & 0.00 & 0.00 & 0.13 \\
\textit{posh} & 0.66 & 0.00 & 0.00 & 0.05\\
\hline
\textit{proprietor} & 0.52 & 0.10 & 0.06 & 0.12 \\
\textit{queue} & 0.63 & 0.10 & 0.08 & 0.08 \\
\hline
\textit{abolish} & 0.80 & 0.22 & 0.23 & 0.28 \\
\textit{bicker} & 0.61 & 0.22 & 0.03 & 0.20 \\
\hline
\textit{isolationist} & 0.79 & 0.32 & 0.17 & 0.30 \\
\textit{justice} & 0.82 & 0.32 & 0.24 & 0.22 \\
\hline
\textit{reminisce} & 0.79 & 0.42 & 0.02 & 0.39 \\
\textit{weeping} & 0.78 & 0.42 & 0.38 & 0.38 \\
\hline
\textit{dictatorship} & 0.88 & 0.68 & 0.48 & 0.53 \\
\textit{totalitarian} & 0.88 & 0.69 & 0.42 & 0.51 \\
\hline
\textit{ten} & 0.93 & 0.77 & 0.57 & 0.69 \\
\textit{twelve} & 0.94 & 0.77 & 0.62 & 0.70 \\
\hline
  \end{tabular}
  \caption{Examples With Different Levels of Overlap, North America Compared to All Other Varieties}
  \label{tab:4a}
\end{table}

Because different vocabulary items are generally learned at different stages of language acquisition, we also include age-of-acquisition ratings for the vocabulary \cite{Kuperman2012}. These ratings are collected via MechanicalTurk but validated against ground-truth age-of-acquisition ratings collected in a laboratory setting. For instance, words like \textit{mom}, \textit{water}, and \textit{yes} are reported to be learned during a child's second year. But words like \textit{constrain}, \textit{confound}, and \textit{thyme} are reported to only be learned at the age of twelve. If more socially-conditioned words are subject to more variation, we might expect, then, that vocabulary learned later in life is subject to more variation as a result. Note that both sets of participant-based ratings (age-of-acquisition and concreteness) depend on inner-circle participants. Thus, these experiments are focused on variation in embedding spaces rather than variation in participant-based lexical features.

The next source of lexical annotations is the \textsc{ucrel} Semantic Analysis system \cite{Piao2015} which provides a high-level semantic domain for each vocabulary item. For example, there are 586 items belonging to the domain \textsc{food and farming} and 761 to the domain \textsc{government and public}. The inventory of semantic domains is shown in Table \ref{tab:4} along with the average concreteness and average age-of-acquisition for each. There is a clear relationship between semantic domain and concreteness: for example, the domain that includes \textsc{psychological states} is highly abstract at 2.4 while the domain that includes \textsc{food and farming} is highly concrete at 4.2. In the same way, some semantic domains are acquired early (like \textsc{names and grammar} at 7.5 years of age) and others much later (like \textsc{science and technology} at 11.4 years of age).

\begin{figure*}[t]
\centering
\includegraphics[width = 380pt]{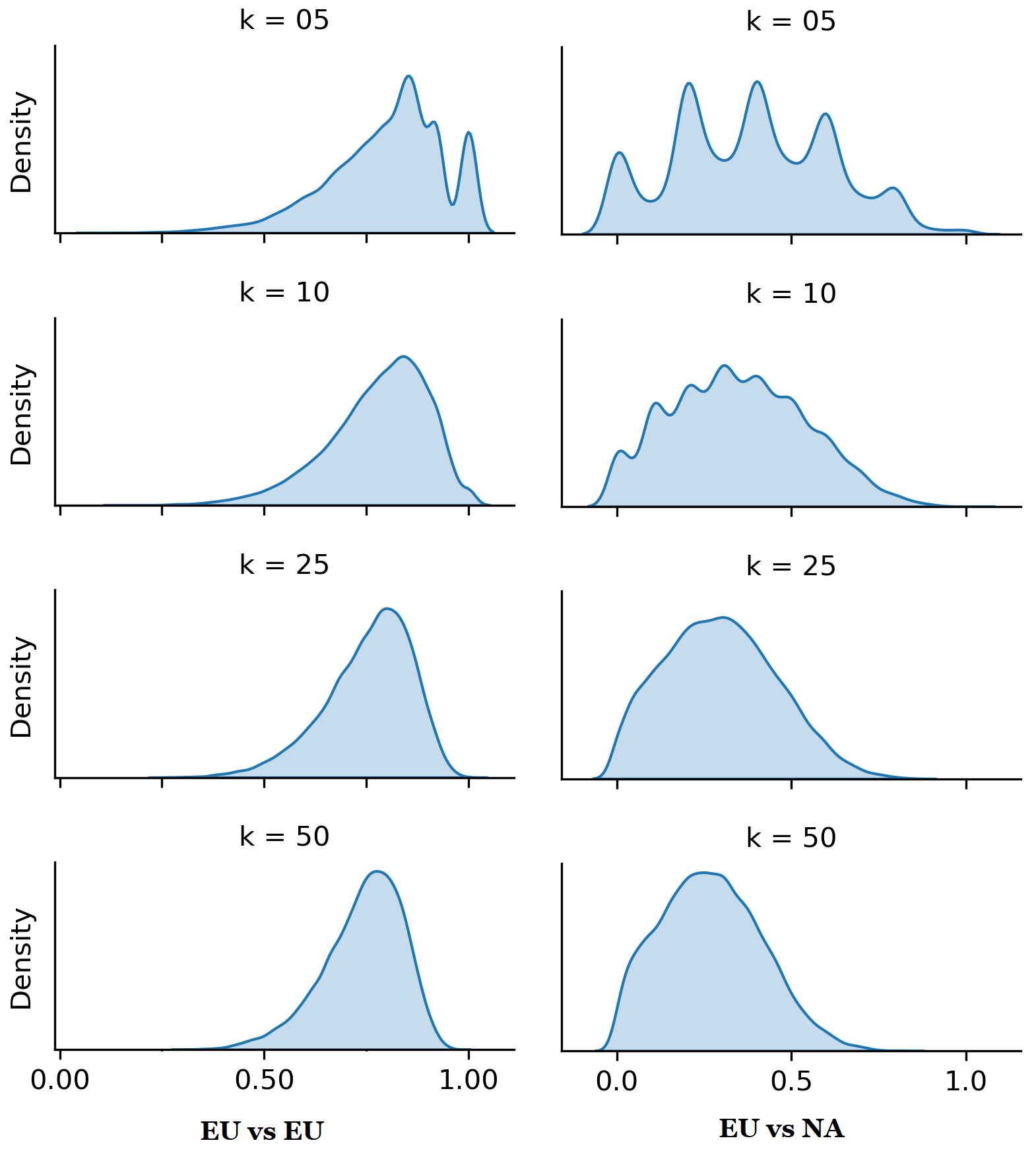}
\caption{Distribution of Overlap Values Across Settings of $k$, for Europe-Europe and Europe-America Comparisons}
\label{fig1}
\end{figure*}

In addition to these participant-based and semantic-based annotations, each lexical item also belongs to a frequency strata. This is calculated using the entire corpus across all regions and reported in occurrences per 1 million words. 

\begin{figure*}[t]
\centering
\includegraphics[width = 450pt]{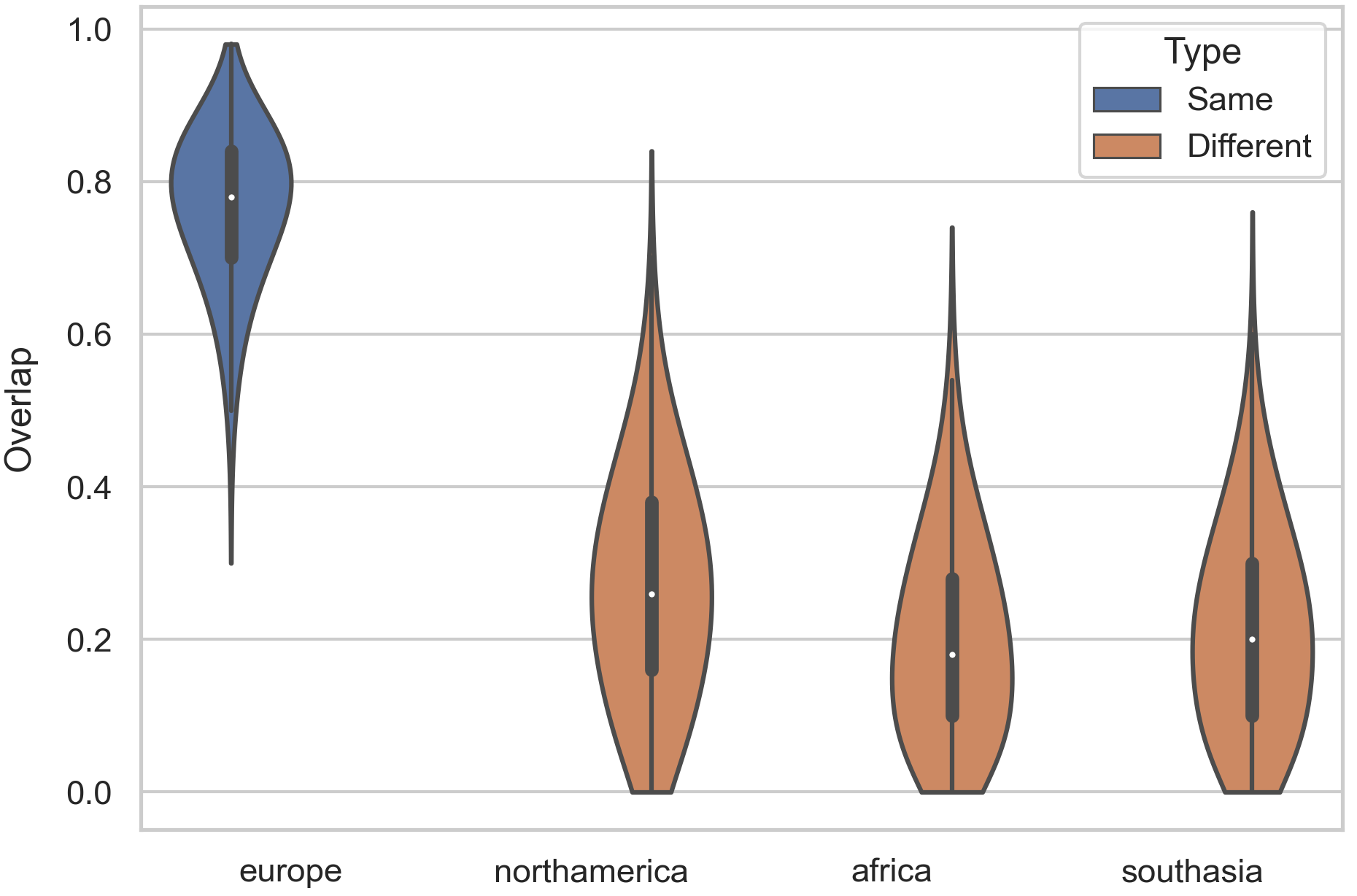}
\caption{Distribution of Within-Dialect vs Between-Dialect Overlap Values for Europe}
\label{fig2}
\end{figure*}

\textbf{Calculating Overlap}. The stability and similarity of word representations are calculated using the overlap of nearest neighbors \cite{Burdick2021}. Given two sets of embeddings (i.e., North America and Europe) we iterate over each word in the lexicon. First, we retrieve the $k$ nearest neighbors using cosine similarity. Second, we calculate the overlap between the two sets of nearest neighbors. For example, if all ten out of ten words appear in both embeddings as nearest neighbors, the overlap is 100\%. If only five words out of ten appear as neighbors, the overlap is 50\% (five shared words out of 10 possible shared words). This provides a word-specific measure of overlap. This method aligns the vocabulary space rather than aligning the embedding spaces; this approach is taken because alignment methods have previously been shown to be unstable \cite{gonen-etal-2020-simple} and thus less suitable for identifying variation across dialects.

A selection of example levels of overlap is shown in Table \ref{tab:4a}, with the North American embeddings compared with all other dialects. The smallest amount of overlap is shown for words like \textit{daft} and \textit{posh} which are used in different senses across these dialects. Culture-specific words like \textit{isolationist} and \textit{justice} provide a mid-level of overlap, with a similar sense but different references across dialects. Finally, a further cultural influence is shown for political words like \textit{dictatorship}, which are more similar in inner-circle dialects than in outer-circle dialects. These examples show the range of overlap levels that are observed.

We measure overlap with values for $k$ of 5, 10, 25, and 50. The distribution of overlap values is shown in Figure \ref{fig1} for the European and North American model (on the right) and for the European and European model (on the left). Thus, the distributions on the right are across dialects and those on the left are within the same dialect. The impact of $k$ is shown in the plots, with $k=5$ at top and $k=50$ at bottom. Smaller values of $k$ lead to ragged distributions simply because the number of possible overlap values is limited. How much impact does the choice of $k$ have on the results? We can see that higher values lead to finer estimates of the distribution of overlap, but overall the values are much the same once $k$ is above 10. For example, there is a significant Pearson correlation between overlap values at $k=25$ and $k=50$ (0.937 on the right and 0.879 on the left, in both cases with $p<0.001$). We use $k=50$ for the rest of the analysis, but the choice of $k$ (above 10) has minimal impact on the results. We also see that the overlap within the same dialect (on the left) is greater than the overlap between different dialects (on the right). The figures for all distributions are available in the supplementary material.\footnote{\href{https://jdunn.name/2023/03/27/variation-and-instability-in-dialect-based-embedding-spaces/}{https://jdunn.name/2023/03/27/variation-and-instability-in-dialect-based-embedding-spaces/}}

\section{Overlap Within vs Between Dialects}

The first experiment evaluates whether the variation between dialects remains meaningful when compared with baseline instability within a single dialect. The overlap measure described above compares the similarity between two sets of embeddings. We visualize the within vs between dialect condition in Figure \ref{fig2} for Europe, with each type of comparison a separate violin plot. In blue we see the within-dialect overlap in which we compare European embeddings to other European embeddings. In orange we see between-dialect overlap for each of the other three regions. There is a clear distinction here between variation within the same dialect (baseline instability) and between different dialects (actual variation).

While we measured overlap between ten unique pairs of embeddings for each condition, this figure shows only the first pair for each. The supplementary materials contain the figures for all comparisons. The conclusion remains the same: the variation in embeddings across dialects is not simply a result of instability alone. There is a clear visual distinction between within-dialect and between-dialect overlap in all cases. We test for significance using a paired t-test: for example, are the values for Europe-Europe comparisons actually different from the values for Europe-Africa comparisons? For each comparison, we randomly choose a single pair of embeddings to test (i.e., so that we compare Europe and Africa only once). In each case the difference is significant with $p<0.001$. 

\begin{table}[h]
\centering
\begin{tabular}{|r|c|c|c|c|}
\hline
~ & \textbf{EU} & \textbf{NA} & \textbf{AF} & \textbf{SA} \\
\hline
\textbf{EU} & \textbf{74.1\%} & 26.8\% & 19.7\% & 21.5\% \\
\textbf{NA} & -- & \textbf{74.0\%} & 20.3\% & 22.0\% \\
\textbf{AF} & -- & -- & \textbf{71.5\%} & 20.7\% \\
\textbf{SA} & -- & -- & -- & \textbf{72.8\%} \\
\hline
  \end{tabular}
  \caption{Bayesian Estimates of Overall Overlap Within and Between Dialects at 95\% Confidence Interval, Across All Comparisons, $k=50$}
  \label{tab:5}
\end{table}

Thus, there is a visually clear and statistically significant difference between baseline instability and variation across dialects. We quantify the magnitude of this difference in Table \ref{tab:5} using a Bayesian estimate of the mean difference across the entire vocabulary and all pairs of embeddings. Within-dialect overlap ranges from 71.5\% to 74.1\%, providing a baseline for instability. Between-dialect overlap ranges from 19.7\% to 26.8\%, showing that the dialect represented by the training corpus has a large influence on downstream embeddings. 

\begin{table*}[t]
\centering
\begin{tabular}{|ll|cc|cc|cc|cc|}
\hline
\multicolumn{2}{|c|}{\textbf{Positive Factors}} & \multicolumn{2}{c|}{\textbf{Europe}} & \multicolumn{2}{c|}{\textbf{N. America}} & \multicolumn{2}{c|}{\textbf{Africa}} & \multicolumn{2}{c|}{\textbf{South Asia}} \\
\multicolumn{2}{|c|}{\textit{Subject to Less Variation}} & \textit{coef.} & \textit{p} & \textit{coef.} & \textit{p} & \textit{coef.} & \textit{p} & \textit{coef.} & \textit{p} \\
\hline
Domain & Body, Individual & 3.97 & 0.000 & 4.36 & 0.000 & 3.82 & 0.000 & 4.56 & 0.000 \\
Domain & Science, Tech & 3.10 & 0.000 & 4.19 & 0.000 & 2.19 & 0.000 & 3.25 & 0.000 \\
Domain & Food, Farming & 2.67 & 0.000 & 2.98 & 0.000 & 1.87 & 0.000 & 3.25 & 0.000 \\
Domain & Emotion & 1.44 & 0.000 & 1.67 & 0.000 & 0.58 & 0.002 & 1.08 & 0.000 \\
Domain & Arts, Crafts & 1.37 & 0.004 & -- & -- & -- & -- & 1.21 & 0.008 \\
Domain & Govt., Public & 1.28 & 0.000 & 1.87 & 0.000 & 1.57 & 0.000 & 1.60 & 0.000 \\
Domain & Entertainment & 0.84 & 0.001 & 0.75 & 0.004 & -- & -- & -- & -- \\
Domain & World, Environ. & -- & -- & 0.91 & 0.003 & -- & -- & 1.22 & 0.000 \\
Domain & Psychological & -- & -- & 0.65 & 0.000 & -- & -- & 0.46 & 0.005 \\
Domain & Social Actions &-- & -- & 0.49 & 0.001 & -- & -- & -- & -- \\
\hline
\textsc{pos} & Verb & 2.58 & 0.000 & 2.48 & 0.000 & 2.71 & 0.000 & 2.26 & 0.000 \\
\textsc{pos} & Function & 12.97 & 0.000 & 10.35 & 0.000 & 12.14 & 0.000 & 10.43 & 0.000 \\
\textsc{pos} & Adverb & 8.83 & 0.000 & 7.23 & 0.000 & 8.47 & 0.000 & 6.55 & 0.000 \\
\hline
\multicolumn{2}{|c|}{\textbf{Negative Factors}} & \multicolumn{2}{c|}{\textbf{Europe}} & \multicolumn{2}{c|}{\textbf{N. America}} & \multicolumn{2}{c|}{\textbf{Africa}} & \multicolumn{2}{c|}{\textbf{South Asia}} \\
\multicolumn{2}{|c|}{\textit{Subject to More Variation}} & \textit{coef.} & \textit{p} & \textit{coef.} & \textit{p} & \textit{coef.} & \textit{p} & \textit{coef.} & \textit{p} \\
\hline
Domain & Communication & -0.59 & 0.002 & -- & -- & -0.77 & 0.000 & -0.59 & 0.001 \\
Domain & Money, Com. & -1.07 & 0.000 & -0.88 & 0.000 & -- & -- & -0.67 & 0.004 \\
Domain & Life, Living & -1.12 & 0.000 & -- & -- & -1.72 & 0.000 & -- & -- \\
Domain & Materials, Objects & -1.14 & 0.000 & -0.91 & 0.000 & -1.42 & 0.000 & -0.63 & 0.000 \\
Domain & Movement, Travel & -1.94 & 0.000 & -1.50 & 0.000 & -2.14 & 0.000 & -1.28 & 0.000 \\
Domain & Housing, Home & -2.19 & 0.000 & -2.38 & 0.000 & -2.39 & 0.000 & -1.54 & 0.000 \\
Domain & Name, Grammar & -2.24 & 0.000 & -- & -- & -2.10 & 0.000 & -- & -- \\
\hline
\textsc{pos} & Names & -5.81 & 0.000 & -6.40 & 0.000 & -4.67 & 0.000 & -6.19 & 0.000 \\
\textsc{pos} & Noun & -- & -- & -0.34 & 0.001 & -- & -- & -0.40 & 0.000 \\
\hline
\hline
\multicolumn{2}{|c|}{\textbf{Scalar Factors}} & \multicolumn{2}{c|}{\textbf{Europe}} & \multicolumn{2}{c|}{\textbf{N. America}} & \multicolumn{2}{c|}{\textbf{Africa}} & \multicolumn{2}{c|}{\textbf{South Asia}} \\
\multicolumn{2}{|c|}{\textit{Lower Ratings=Less Variation}} & \textit{coef.} & \textit{p} & \textit{coef.} & \textit{p} & \textit{coef.} & \textit{p} & \textit{coef.} & \textit{p} \\
\hline
Empirical & AoA & -0.54 & 0.000 & -0.53 & 0.000 & -0.56 & 0.000 & -0.48 & 0.000 \\
Empirical & Concreteness & -1.66 & 0.000 & -1.72 & 0.000 & -1.69 & 0.000 & -1.74 & 0.000 \\
\hline
  \end{tabular}
  \caption{Coefficients and P-Values from a Linear Mixed Effects Regression Model Using the Mean Overlap Across Dialects as the Dependent Variable. Non-Significant Effects are Not Shown.}
  \label{tab:6}
\end{table*}

There is a slight effect for inner-circle and outer-circle dialects: North America \textsc{(na)} and Europe \textsc{(eu)} are more similar to one another than to Africa \textsc{(af)} or South Asia \textsc{(sa)}. Compared to the distinction between variation and baseline instability, however, this effect is relatively minor. The outer-circle varieties also have slightly lower stability than the inner-circle varieties.

\section{Lexical Factors}

We have shown that there is a significant difference in embedding spaces depending on the dialect represented in the training data, a difference that is much greater than baseline instability within dialects (as simulated by shuffling and retraining on the same corpora). This section explores dialectal variation in embedding spaces further by focusing on the impact of the lexical factors described in Section 5. We ask whether this kind of variation is distributed equally across the lexicon or whether it is concentrated in particular types of vocabulary.

We model the relationship between lexical attributes and overlap using a linear mixed effects regression model, with one model for each dialect. Within each model, the region of comparison is a fixed effect: for example, we model variation within the European embeddings using their overlap with North America, Africa, and South Asia as fixed effects. For random effects we include all lexical attributes. We represent each region using the average overlap across all ten pairs of embeddings, using $k=50$ as before. The means of different regions are independent in the sense that each vocabulary item is modelled independently from corpora representing that region.

The coefficients and p-values for each lexical attribute are shown in Table \ref{tab:6} for all attributes that are significant for at least one dialect ($p<0.01$). Positive categorical factors are shown above and negative factors below. Columns show results from the four dialect-specific models. While some factors are significant in one dialect but not another, no attributes have opposite effects across dialects (i.e., indicate more variation in one dialect but less variation in another).

Within semantic domains, vocabulary involving \textsc{body and individual} (e.g., \textit{pain} and \textit{ache}) are more stable across dialects, as are \textsc{food and farming} (e.g., \textit{celery} and \textit{sushi}) and \textsc{science and technology} (e.g., \textit{biologist} and \textit{geologist}). These terms are less socially-conditioned in the sense that they refer to tangible objects or to specially-defined fields (like biology) that transcend cultural boundaries. On the other hand, vocabulary from semantic domains \textsc{home and housing} (e.g., \textit{guest} or \textit{pew}), \textsc{movement and travel} (e.g., \textit{turnpike} or \textit{curbside}), and \textsc{names and grammar} (e.g., \textit{northwestern} or \textit{roman}) are subject to more variation. These words are more socially-conditioned in the sense that they presume socially-defined concepts: a \textit{guest} requires a definition of family units and a \textit{pew} is a part of the concept \textsc{church}. Within parts-of-speech, function words (e.g., \textit{of} or \textit{and}) and adverbs (e.g., \textit{hardly} and \textit{exactly}) are much more stable. And named entities (e.g., \textit{Flint}) are much less stable. Verbs are more important to the model than nouns.

Of the three scalar attributes, frequency has a significant effect but the coefficient is so small it is negligible. Concreteness is significant in every region, with more abstract words (e.g., \textit{surreal} and \textit{sanctimonious}) being more stable while more concrete words (e.g., \textit{cookie} and \textit{bug}) are less stable. In this case, the specific instances (the referents) of these more concrete terms are likely to be quite different across dialects (\textit{cookies} are different in different places). Age-of-acquisition is significant in three out of four regions, but it has only a relatively small effect, with words acquired at a younger age being more stable. For instance, \textit{mother} and \textit{grandmother} (learned at age 2) are quite stable while \textit{ethos} and \textit{polarization} (learned at age 15) are subject to variation. The full regression results and the stability/variability values for the entire lexicon are available in the supplementary materials.\footnote{\href{https://jdunn.name/2023/03/27/variation-and-instability-in-dialect-based-embedding-spaces/}{https://jdunn.name/2023/03/27/variation-and-instability-in-dialect-based-embedding-spaces/}}

\section{Discussion and Conclusions}

These experiments have shown that embedding spaces are subject to variation according to the dialect represented by the training data. This variation is significantly greater than noise caused by baseline instability in the embeddings themselves. This finding confirms the importance of regional dialects in \textsc{nlp}: while previous work has shown the impact of dialect on lexical and syntactic representations, this paper shows that such variation also extends to semantic representations.

Previous work has focused on distinguishing between dialects or on directly modelling variation over space and time. This paper has taken a different approach by training otherwise comparable models on corpora representing different dialects, controlling for other sources of variation like parameter settings and random seeds. The results show that the dialects represented in the training context have significant downstream impacts on common semantic representations (embeddings). These findings raise important questions for future work. First, is the influence of dialect consistent across languages or is this a result of the colonial history of a few languages like English? Second, do contextual embeddings also manifest this type of variation or is it confined to non-contextual skip-gram embeddings? Third, would a larger inventory of dialect-specific embeddings change the distribution of variation within the lexicon or is this a stable effect? Regardless of such further questions, these experiments show that dialect has a downstream effect on semantic representations, expanding previous work on lexical and syntactic representations.

% Entries for the entire Anthology, followed by custom entries
\bibliography{aacl}
\bibliographystyle{acl_natbib}

\end{document}